\documentclass[10pt,a4paper]{article}

\usepackage{amsmath,amsgen,latexsym}
\usepackage{amstext,amssymb,amsfonts,latexsym}
\usepackage{theorem}
\usepackage{pifont}
\usepackage[dvips]{graphics,epsfig}

 \newcommand{\bs}{\bigskip}
 
 \newcommand{\n}{\noindent}
 \newcommand{\s}{\smallskip}
 \newcommand{\hs}[1]{\hspace*{ #1 mm}}
 \newcommand{\vs}[1]{\vspace*{ #1 mm}}

 \newcommand{\setempty}{\mathrm{\O}}
 \newcommand{\real}{\mathbb{R}}
 \newcommand{\nat}{\mathbb{N}}
 
 \newcommand{\integer}{\mathbb{Z}}
 \newcommand{\rational}{\mathbb{Q}}

 \newcommand{\ie}{\textrm{i.e.},\hspace*{2mm}}
 \newcommand{\eg}{\textrm{e.g.},\hspace*{2mm}}

 \newcommand{\FF}{{\cal F}}

 \newcommand{\LL}{{\cal L}}

 \newcommand{\PP}{{\cal P}}

 \newcommand{\np}{\mathrm{NP}}

 \newcommand{\fp}{\mathrm{FP}}

\theoremstyle{plain}
\theoremheaderfont{\bfseries}
 \newtheorem{theorem}{Theorem}[section]
 \newtheorem{lemma}[theorem]{Lemma}
 \newtheorem{proposition}[theorem]{Proposition}

 {\theorembodyfont{\rmfamily}
  \newtheorem{definition}[theorem]{Definition}}
 {\theorembodyfont{\rmfamily} }

 \newenvironment{proofsketch}{\par \noindent
            {\bf Proof Sketch. \hs{1}}}{\hfill$\Box$ \vspace*{3mm}}

 \newcommand{\pair}[1]{\langle #1 \rangle}

\newcommand{\ignore}[1]{}

\newcommand{\cent}{{|}\!\!\mathrm{c}}
\newcommand{\dollar}{\$}

\newcommand{\supp}{\mathrm{Supp}}

\newcommand{\cut}{\mathrm{Cut}}
\newcommand{\ball}{\mathrm{Ball}}
\newcommand{\AFreduces}{\leq_{\mathrm{AF}}}
\newcommand{\APFreduces}{\leq_{\mathrm{APF}}}
\newcommand{\fuzzypa}{\mathrm{Fuzzy}\mbox{-}\mathrm{PA}}
\newcommand{\fuzzynpa}{\mathrm{Fuzzy}\mbox{-}\mathrm{NPA}}

\newcommand{\fuzzyfpa}{\mathrm{Fuzzy}\mbox{-}\mathrm{FPA}}

\newcommand{\fuzzynpao}{\mathrm{Fuzzy}\mbox{-}\mathrm{NPAO}}

\begin{document}
\pagestyle{plain}
\setcounter{page}{1}

\begin{center}
{\Large {\bf The World of Combinatorial Fuzzy Problems and the \s\\
Efficiency of Fuzzy Approximation Algorithms}}\footnote{This extended abstract appeared in the Proceedings of the Joint 7th International Conference on Soft Computing and Intelligent Systems (SCIS 2014) and 15th International Symposium on Advanced Intelligent Systems (ISIS 2014), December 3--6, 2014, Institute of Electrical and Electronics Engineers (IEEE), pp. 29--35, 2014.} \bs\\
{\sc Tomoyuki Yamakami}\footnote{Present Affiliation: Department of Information Science, University of Fukui, 3-9-1 Bunkyo, Fukui 910-8507, Japan} \bs\\
\end{center}

\begin{quote}
{\small \n{\bf Abstract:}
We re-examine a practical aspect of combinatorial fuzzy problems of various types, including search, counting, optimization, and decision problems. We are focused only on those fuzzy problems that take series of fuzzy input objects and produce fuzzy values. To solve such problems efficiently, we design fast fuzzy algorithms, which are modeled by polynomial-time deterministic
fuzzy Turing machines equipped with read-only auxiliary tapes and
write-only output tapes and also modeled by polynomial-size fuzzy circuits composed of fuzzy gates. We also introduce fuzzy proof verification systems to model the fuzzification of nondeterminism.
Those models help us identify four complexity classes: Fuzzy-FPA of fuzzy functions, Fuzzy-PA and Fuzzy-NPA of fuzzy decision problems, and Fuzzy-NPAO of fuzzy optimization problems.
Based on a relative approximation scheme targeting fuzzy membership degree, we formulate two notions of ``reducibility'' in order to compare the computational complexity of two fuzzy problems. These reducibility notions  make it possible to locate the most difficult fuzzy problems in Fuzzy-NPA and in Fuzzy-NPAO.
}
\end{quote}

\sloppy
\section{Background and Results}

Our purpose is to (1) make a theoretical groundwork necessary to carry out practical analyses of ``generalized'' fuzzy problems that have naturally arisen in industrial applications and (2) lay out a theoretical framework for ``generalized'' fuzzy algorithms that efficiently solve those ``generalized''  fuzzy problems.

\subsection{Practical Realm of Fuzzy Problems}\label{sec:fuzzy-problems}

In real-life situations, many objects naturally embody certain degrees of {\em fuzziness}, which can be in general expressed in terms of the uncertainty, ambiguity, vagueness, or imprecision of the objects. Based on a non-standard logic, Zadeh \cite{Zad65} tried to capture this fuzziness mathematically by formulating a basic concept of {\em fuzzy set}. In his theory, an ordinary object without any fuzziness is distinctly called as a ``crisp'' object, whereas a fuzzy object is a combination of crisp object and its membership degree (or certainty degree) that indicates intuitively how likely the crisp object actually exists.
Since its introduction, fuzzy theory has found numerous applications from digital image processing to voice recognition, to telecommunication, further to a field of medicine and agriculture (see, \eg \cite{AKLM01,Han10}). To handle a wide variety of practical fuzzy problems, we still need to lay out a groundwork in developing a general, coherent theory of fuzzy problems and fuzzy algorithms that efficiently solve them.

In the rest of this paper, we wish to limit our interest within combinatorial problems, which include search, counting, optimization, and decision problems. Let us first recall that, in an ordinary theory, those problems are viewed as functions that map every input instance to its desirable solution (or solutions).
In the past literature, there have been numerous ways to fuzzificate ordinary combinatorial problems and those fuzzification methods significantly vary, depending on target areas of interests.
We wish to re-examine a {\em fuzzification} of those problems to develop our general framework.

In many abstract treatments of fuzzy problems, input instances are merely  pairs of  crisp objects and their membership degrees, instead of more general ``fuzzy objects.'' To expand a scape of fuzzy-logic applications, it is more desirable to deal with the case where an input instance is a series of admissible fuzzy objects (or fuzzy data) and an output could be a desired ``fuzzy object'' as a solution derived from the given input. Formally, we define our ``fuzzy problem'' as a mapping from each series of fuzzy objects to another fuzzy object; thus, the term ``fuzzy problem'' becomes a synonym of ``fuzzy functions.''
To describe those fuzzy problems, we need to specify two items: {\em fuzzy (input)  instance} and {\em fuzzy output}. For technical reason, we assume that the support of any fuzzy instance is finite, where  the {\em support} of a fuzzy object over universe $U$ is
an ordinary set of elements in $U$ having  positive  membership degrees.
To distinguish our fuzzy problems from the aforementioned conventional fuzzy problems, we sometimes emphasize the generosity of our fuzzy problems by calling them {\em generic fuzzy problems}.

\s
\n\hs{1}{\sc (Generic) Fuzzy Problem} $P$: \s\\
\n\hs{1}$\bullet$\hs{2} {\sc Fuzzy instance:} a series  $(s_1,s_2,\ldots,s_k)$ of admissible fuzzy objects having finite supports. \\
\n\hs{1}$\bullet$\hs{2} {\sc Fuzzy output:} an admissible fuzzy object as a solution to this input $(s_1,s_2,\ldots,s_k)$.

\s

Due to the page limit,  we will discuss in later sections only two kinds of (generic) fuzzy problems: fuzzy decision problems and fuzzy optimization problems.
{\em Fuzzy decision problems} are treated as a special case of the fuzzy problems whose outputs are limited to fuzzy subsets of $\{0,1\}$. This gives rise to a new realm of fuzzy languages, each of which indicates input's ``acceptance'' and ``rejection'' with separate precision degrees, whereas conventional ``fuzzy languages'' are defined simply as fuzzy subsets of $\Sigma^*$ for underlying alphabet $\Sigma$.

Fuzzy optimization problems are also a variant of (generic) fuzzy problems,
which  in general have the form $(I,SOL,m,goal)$, where $I$ is a finite set of fuzzy (admissible) input instances, $SOL$ is a fuzzy function listing all fuzzy solutions, $m$ is a fuzzy measure (or objective) function from $I\circ SOL$ to natural numbers, and $goal\in\{\max,\min\}$. Here $I\circ SOL$ means the set  $\{(s,t)\mid s\in I,t\in SOL(s)\}$. Let $m^*(x)$ denote the optimal value $m(x,y)$ over all solutions $y\in SOL(x)$.

\subsection{Practical Realm of Fuzzy Algorithms}

In Section \ref{sec:fuzzy-problems}, we have discussed our (generic) fuzzy problems. To solve those problems efficiently, we need to clarify how we model efficient fuzzy algorithms  so that we can explore computational  complexity issues of the fuzzy problems in a wider, generic framework.

We intend to model fuzzy algorithms by extending the existing realms of ``sequential computation'' and ``parallel computation.''
Firstly, we will model our fuzzy algorithms by refining the existing notion of fuzzy Turing machines, which were considered by Zadeh \cite{Zad68}, Lee and Zadeh \cite{LZ69}, and Santos \cite{San70} as a fuzzification of ordinary nondeterministic  Turing machines (or NTMs, in short) .
In their formulations, deterministic fuzzy computation inherently embodies ordinary nondeterminism, and therefore it is not surprising to know that, under a certain suitable fuzzification of ordinary instances, fuzzy algorithms solve NP-complete problems in polynomial time (cf. Proposition \ref{NP-vs-Fuzzy-PA}).

To cope with a wide range of practical fuzzy problems, Doostfatemeh and Kremer \cite{DK05} suggested how to expand the then-existing models of fuzzy algorithms by supplementing extra ``safe'' auxiliary operators to tune up the behaviors of those algorithms. Based on their spirit, we give a new formulation of {\em deterministic fuzzy Turing machines} (or DFTMs).

Another natural model that can represent fuzzy algorithms is {\em fuzzy (logic) circuits} (\eg  \cite{GV89,Hir94}). In particular, we consider fuzzy circuits in which fuzzy gates are {\em layered} level by level.
Such a fuzzy circuit takes a series of fuzzy input bits and proceeds by applying fuzzy gates at each level. This mechanism can realize fuzzified ``parallel computation.''
We prove that families of polynomial-size fuzzy circuits are equivalent in computational power to polynomial-time DFTMs.

The minimum amount of computational resources necessary to solve given problems is of great concern from a practical viewpoint. As in ordinary computational complexity theory, we are also interested in fuzzy algorithms running in polynomial time.
For the purpose of solving a wider scope of fuzzy problems, we look for an approximation of the outcomes of fuzzy algorithms.
We therefore define $\fuzzyfpa(\gamma)$ as the class of fuzzy functions (or equivalently, fuzzy problems having output values) computed approximately with relative closeness of $\gamma$ by DFTMs in polynomial time. Restricted to fuzzy decision problems, we also obtain $\fuzzypa(\gamma)$ as a natural fuzzification of the complexity class P.

Nondeterminism has been modeled as a proof-verification process, in which, for positive instances, there exists a ``proof'' (\ie necessary key information) for which a ``verifier'' can easily confirm its validity and, for negative instances, the verifier refutes any proof provided to him as invalid.
We recognize a naturally-induced class $\fuzzynpa(\gamma)$ of fuzzy decision problems that are solved approximately with relative closeness $\gamma$
by such fuzzy proof verification systems (or FPVS's, in short) in polynomial-time.

To discuss the computational complexity of fuzzy problems, we introduce a fundamental notion of {\em polynomial-time approximate fuzzy reducibility} (abbreviated as AF-reducibility) between two fuzzy problems.
Such a reducibility notion among fuzzy decision problems makes it possible to designate ``complete'' problems, which indicate the most difficult problems in a given class of problems to solve in polynomial time, as for the notion of NP-complete problems (see \cite{GJ79}).
We prove
the existence of complete problems in $\fuzzynpa(1)$ under AF-reductions.

For fuzzy optimization problems, similarly to $\fuzzynpa(\gamma)$, we define $\fuzzynpao(\gamma)$, which is composed of all fuzzy optimization problems characterized by fuzzy functions in $\fuzzyfpa(\gamma)$.
Between two fuzzy optimization problems, we also introduce a notion of {\em polynomial-time approximation-preserving fuzzy reducibility} (or APF-reducibility) and prove that $\fuzzynpao(1)$ contains complete problems under APF-reductions.

All omitted or abridged proofs will appear in a complete version of this extended abstract.

\section{Basic Notions and Notations}

The notation $\integer$ (resp., $\rational$, $\real$) denotes the set of all integers (resp., rational numbers, real numbers). We use $\nat$ to denote the set of all natural numbers (\ie non-negative integers) and we set $\nat^{+} =\nat-\{0\}$.
For two integers $n,m$ with $n\leq m$, $[n,m]_{\integer}$ stands for the {\em integer interval} $\{n,n+1,n+2,\ldots,m\}$. For simplicity, we write $[n]$ to denote the interval $[1,n]_{\integer}$ whenever $n\geq1$.  The notation $[0,1]^*$ (resp., $[0,1]^{+}$) denotes the union of $[0,1]^k$ for any constant $k\in\nat$ (resp., $k\in\nat^{+}$).
Given any propositional formula $F$, we set $[F]$ to be $1$ if $F$ is true, and $0$ otherwise. For example, $[x=y]$ equals $1$ if and only if $x=y$.
For any set $A$, $\PP(A)$ denotes the {\em power set} of $A$.

An {\em alphabet} $\Sigma$ is a finite nonempty set of ``symbols'' and a {\em string} over $\Sigma$ is a finite sequence of symbols taken from $\Sigma$; in particular, the {\em empty string} is always denoted $\lambda$. Let $\Sigma^*$ denote the set of all strings over $\Sigma$.
For  any logical proposition (or logical statement) $P$, the notation $[P]$ denotes $1$ if $P$ is true, and $0$ otherwise. For example, we have  $[x=y]=1$ if $x$ equals $y$, and $[x=y]=0$ if $x$ is different from $y$.


Due to the page limit, we assume that the reader is familiar with basic concepts of fuzzy sets.
Although a {\em fuzzy (sub)set} $A$ of universe $U$ is a map from $U$ to $[0,1]$, it is also viewed as an ordinary set that is composed of pairs of the form
$(x,\gamma)$, where $x\in U$ and $\gamma\in[0,1]$; that is, $(x,\gamma)\in A$ iff $\gamma=A(x)$.
To simplify the descriptions of fuzzy sets in the rest of the paper, we  will take those two different viewpoints interchangeably. For convenience, we write $\FF(U)$ for the collection of all fuzzy subsets of $U$. Moreover, we use the notations $core(s)$ for the {\em core} of fuzzy set $s$ and $\supp(s)$ for the {\em support} of $s$.

\begin{definition}[fuzzy function]
Given two universes $U_1$ and $U_2$, let $A\subseteq\FF(U_1)$ and $B\subseteq\FF(U_2)$. A {\em (generic) fuzzy function $f$ from $A$ to $B$}, written as $f:A\rightarrow B$, satisfies that, for every $s\in A$, $f(s)$ is a fuzzy subset of $U_2$ in $B$.
\end{definition}

\n{\bf Remark:}
It may be possible to expand our definition of fuzzy sets by replacing the unit interval $[0,1]$ with an arbitrary complete lattice and expand our results further. However, we leave such a challenging task to the interested  reader.

\section{Combinatorial Fuzzy Problems}\label{sec:fuzzy-problem}

In pursuit of simplicity, we limit our attention within ``combinatorial problems'' that inherently embody fuzziness. Now, we want to explain what ``fuzzy problems'' are. As described in Section \ref{sec:fuzzy-problems},
fuzzy problems with which we deal in this paper take series of fuzzy instances (such as fuzzy graphs, fuzzy numbers, fuzzy functions, etc.) as inputs and produce certain fuzzy objects as outputs. We treat such a problem as a mapping from a set of fuzzy objects (or fuzzy instances) to another fuzzy object.

As customarily in ordinary computational complexity theory,
we will deal only with {\em discrete} objects as our input
instances.
For a general treatment of such discrete fuzzy instances, it is useful to assume a suitable encoding of those instances to a certain fixed object.
In a way similar to defining ``strings over alphabet'' in ordinary language theory, we will introduce an exquisite notion of ``fuzzy string,'' which expresses a certain degree of imprecision, uncertainty, or incompleteness of string information.
Our definition of fuzzy string is inspired by the notion of {\em $LR$-shape fuzzy numbers} (see, \eg \cite{Han10}) and is closely related to ``discretized'' fuzzy numbers.

A {\em fuzzy string quantity $s$ over $\Sigma$} is an element of $\FF(\Sigma^*)$, namely, a fuzzy subset of $\Sigma^*$. We call each value $s(x)$ a {\em precision degree of $x$} instead of ``membership degree.''
A {\em fuzzy string over $\Sigma$} (with respect to a suitably chosen
distance measure $d$, \eg the Hamming distance between two crisp strings) is a fuzzy string quantity $s$ over $\Sigma$ that satisfies the following condition: there exist a crisp string $x_0\in\Sigma^*$ and a crisp function $\eta:[0,1]\rightarrow[0,1]$ (which may depend on $x_0$) such that (1) $\supp(s)$ is  a finite set, (2) $\eta$ is strictly decreasing function with $\eta(0)=1$ and $\eta(1)=0$, and (3) $\cut_{\gamma}(s)\subseteq \ball_{d}(x_0,\eta(\gamma))$ holds for any real number $\gamma\in[0,1]$, where $\ball_{d}(x,\gamma)=\{z\in\Sigma^*\mid  d(x,z)  \leq \gamma\}$. This special string $x_0$ is called a {\em target} of the fuzzy string $s$. Let $F\Sigma^*$ denote a collection of all (possible) fuzzy strings over $\Sigma$.
It holds that $core(s)=x_0$ by Conditions (2) and (4). Note that a fuzzy string is a convex fuzzy set. As usual in fuzzy theory, a fuzzy string $s$ can be seen as the set $\{(x,s(x))\mid x\in\Sigma^*\}$. We also follow a standard convention that, when a fuzzy string $s$ is crisp with a target $x_0$ (\ie $s(x_0)=1$ and $s(x)=0$ for any $x\in\Sigma^*-\{x_0\}$), we identify $s$ with $x_0$. This helps us treat all crisp strings as a special case of fuzzy strings.


Henceforth, we will focus our attention only on any subset $F\Theta$ of $F\Sigma^*$ for a certain alphabet $\Sigma$ and a certain distance measure $d$. A {\em (generic) fuzzy problem} stated in Section \ref{sec:fuzzy-problems} is now rephrased as follows.
Fix our alphabet $\Sigma_1,\Sigma_2$ and consider two sets $F\Theta_1\subseteq F\Sigma_1^*$ and $F\Theta_2\subseteq F\Sigma_2^*$.

\s
\n\hs{1}{\sc (Generic) Fuzzy Problem} $L$ w.r.t. $(F\Theta_1,F\Theta_2)$: \s\\
\n\hs{1}$\bullet$\hs{2} {\sc Fuzzy instance:} a series $(s_1,s_2,\ldots,s_k)$ of fuzzy strings in  $F\Theta_1$. \\
\n\hs{1}$\bullet$\hs{2} {\sc Fuzzy Output:} a fuzzy string in $F\Theta_2$.

\s
\n In particular, we will study two types of fuzzy problems: fuzzy decision problems and fuzzy optimization problems, defined in Section \ref{sec:fuzzy-problems}.

\section{Models for Efficient Fuzzy Algorithms}\label{sec:comp-model}

To solve the fuzzy problems described in Section \ref{sec:fuzzy-problem}, we want to design efficient fuzzy algorithms.

\subsection{Auxiliary Operators for Practical Algorithms}

The existing formulation of fuzzy Turing machine seems too restrictive to apply to many real-life fuzzy problems.  For a wider range of practical applications of fuzzy machines,
Doostfatemeh and Kremer \cite{DK05} proposed a rigorous use of ``auxiliary functions'' to a design of fuzzy algorithms (actually, fuzzy finite automata in their case).
For the purpose of intended practical applications, we adopt their
idea and introduce four auxiliary operators $(\mu_1,\mu_2,\mu_3,\xi)$, where $\mu_1:[0,1]^2\rightarrow[0,1]$, $\mu_2:[0,1]^{+}\rightarrow[0,1]$, $\mu_3:[0,1]^{+}\rightarrow[0,1]$, and $\xi:[0,1]^*\rightarrow[0,1]$. The use of them will be clarified in Section \ref{sec:DFTM}.

A tuple $(\mu_1,\mu_2,\mu_3,\xi)$ of auxiliary operators is said to be {\em safe} if those functions satisfy the following conditions: for any finite nonempty ordered set $A$ and any finite ordered set $B$,

\n\hs{1}
(1)\hs{2} $\mu_1(\alpha,\alpha)=\alpha$ for any $\alpha\in[0,1]$, \\
\n\hs{1}
(2)\hs{2} $\mu_2(\{\alpha_r\}_{r\in A})=\alpha$ if $\alpha_r=\alpha$ for all $r\in A$, \\
\n\hs{1}
(3)\hs{2} $\mu_3(\{\alpha_r\}_{r\in A})=\alpha$ if $\alpha_r=\alpha$ for all $r\in A$, and \\
\n\hs{1}
(4)\hs{2} $\xi(\setempty)=0$ and $\xi(\{\alpha_r\}_{r\in B})=\alpha$ if $\alpha_r=\alpha$ for $r\in B$.

As a concrete example, we present a {\em standard} safe tuple $\Xi=(\mu_1,\mu_2,\mu_3,\xi)$, which we will use later.
Let $\wedge$ and $\vee$ be any t-norms. We set $\mu_1(\alpha,\beta)=\alpha\wedge \beta$.  For any finite nonempty ordered set $A$, let $\mu_2(\{\alpha_r\}_{r\in A})= \bigvee_{r\in A}\alpha_r$ and $\mu_3(\{\alpha_r\}_{r\in A}) = \bigvee_{r\in A}\alpha_r$. Moreover, for any finite ordered set $B$, let  $\xi(\setempty)=0$ and $\xi(\{\alpha_r\}_{r\in B})=\bigvee_{r\in B}\alpha_r$.

\subsection{Deterministic Fuzzy Turing Machines}\label{sec:DFTM}

In order to describe a fuzzified ``sequential computation'' within our  framework, we use a mathematical model of {\em deterministic fuzzy Turing machines equipped with read-only auxiliary-input tapes,  rewritable input/work tapes,  and write-only output tapes} supported by safe tuples of auxiliary operators. This model is considered as
an extension of ordinary nondeterministic Turing machines.

Let $\Delta$ and $\Gamma$ be an {\em input alphabet} and an {\em output alphabet} including a designated blank symbol $\#$. To operate
a machine, we need two more tape alphabets $\Sigma_1$ (with $\cent,\dollar,\#\in\Sigma_1$) and $\Sigma_2$ ($=\{\cent,\dollar,1,\#\}$)
for the machine's internal use.
We fix $F\Theta_1\subseteq \FF(\Delta^*)$ and $F\Theta_2\subseteq \FF(\Gamma^*)$ and also fix a safe tuple $\Xi=(\mu_1,\mu_2,\mu_3,\xi)$ of auxiliary operators.
A {\em deterministic fuzzy Turing machine with a write-only output tape}
(abbreviated as DFTM) is a triplet $\pair{M,\Xi,I}$, where $M$ is of the form $(Q,\Delta,\Sigma_1,\Sigma_2,\Gamma,\delta,q_0,F)$ and $I$ is a target set pair $(F\Theta_1,F\Theta_2)$.
Moreover, $Q$ is a finite nonempty set of the machine's (inner) states and  $\delta$ is a {\em fuzzy transition function} mapping from $(Q-F)\times \Sigma_1\times \Sigma_2$ to $\FF(Q\times \Sigma_1 \times\Gamma\times D_1\times D_2)$ with $D_i=\{0,\pm1\}$, $q_0$ ($\in Q$) is the initial state, and
$F$ ($\subseteq Q$) is a set of {\em halting states}. Whenever $M$ enters a halting state, since $\delta$ is no longer applicable, $M$ is considered to {\em  halt}.

Since $\Xi$ and $I$ are fixed, we often refer to $M$ as a DFTM  as long as
it is clear from the context. Graphically speaking, $M$ works as follows.
A {\em fuzzy input instance} $s$ to $M$ is a fuzzy object in $F\Theta_1$. Let $\ell(s)=\max_{x\in \supp(s)}\{|x|\}$ be the {\em length} of $s$.
The DFTM $M$ takes $s$ and eventually produces another fuzzy subset $M(s)$ of $\FF\Theta_2$ as its output.
In an initial setup process, $M$ receives $s$,
it automatically generates {\em in parallel} all crisp strings $x\in \supp(s)$, and writes down each $x$, surrounded by two endmarkers $\cent$ and $\dollar$, on the input/work tape and the other tape cells initially hold the blank symbol $\#$.
Moreover, $M$ generates $\cent 1^{\ell(s)}\dollar$ on the read-only auxiliary input tape. All tape cells are indexed by integers and all tape heads are initially positioned at the $0$th  cell (where $\cent$ must be written on the first two tapes).
An inner state of $M$ is $q_0$.

Assume that, at an arbitrary moment, the tape head of $M$ scans symbol $\sigma_1$ on the input/work tape and $\sigma_2$ on the read-only input tape
in inner state $q$ and the third tape head is situated at the first (\ie the leftmost) blank cell.
When $\delta$ is applied, $M$ overwrites $\sigma_1$ with $\tau$, enters state $p$, move the first two tape heads in directions $d_1$ and $d_2$, and writes symbol $\eta$ on the output tape, together with {\em possibility degree} $\delta(q,\sigma_1,\sigma_2)(p,\tau,\eta,d_1,d_2)$.
If $\eta\neq\lambda$, then the output tape head must move to the right; otherwise, it stays still.

Let us explain how a DFTM generates a ``fuzzy computation,'' composed of  a set of fuzzy computation paths (\ie series of configurations). A {\em (global) configuration} is a string of the form $uqv\natural r \natural w$ with $u,v\in \Sigma_1^*$, $q\in Q$, $r\in[0,\ell(s)]_{\integer}$,
and $w\in\Gamma^*$, where $\natural$ is a designated separator, and $uqv\natural r\natural w$ indicates
that an input/work tape content is $uv$, the second tape head scans cell $r$, an output tape content is $w$,  an inner state is $q$, and the first and third tape heads are respectively scanning the leftmost symbol of $v$ and the first blank symbol to the right of $w$.
Let  $CONF_{M}$ denote the set of all (possible) configurations of $M$, namely, $\{uqv\natural r\natural w \mid u,v\in \Sigma_1^*,q\in Q, r\in[0,\ell(s)]_{\integer}, w\in\Gamma^*\}$.
A {\em final configuration} is a configuration of the form $uqv\natural r \natural w$ with $q\in F$ in $CONF_{M}$.
Each element of $\FF(CONF_{M})$ is called a {\em fuzzy instance description} or a {\em fuzzy configuration}.

Let us describe how a ``time evolution'' of fuzzy configurations takes place. To express a fuzzy configuration at time $i$, we inductively introduce a fuzzy set $conf_i$ ($i\in\nat$) as follows. Let $conf_0$ be a {\em fuzzy configuration at time $0$} defined as
\[
conf_0(upv\natural r\natural w)= \sum_{x\in \supp(s)} s(x)\cdot [upv\natural r \natural w = q_0\cent x\dollar\natural 0 \natural \lambda]
\]
for any configuration $upv\natural r\natural w$ of $M$.
Next, $conf_{i+1}$ is a {\em fuzzy configuration at time $i+1$} whose possibility degree $conf_{i+1}(u'pv'\natural r \natural w')$ is defined as follows. Taking pre-determined orderings on $(Q-F)\times \Sigma_1\times \Sigma_2$ and $D_1\times D_2$ and also assuming that $u'pv'\natural w'$ has the form $u\tau_1p\tau_2\tau_3 v\natural r \natural w\eta$ with $p\in Q-F$ and , let $conf_{i+1}(u'pv'\natural r\natural w')$ be
\[
  \mu_2(\{\ell_{d_1,d_2}(q,\sigma_1,\sigma_2)\}_{(d_1,d_2)\in D_1\times D_2, (q,\sigma_1,\sigma_2)\in (Q-F)\times \Sigma_1\times \Sigma_2}),
\]
where $\ell_{d_1,d_2}(q,\sigma_1,\sigma_2)$ represents the following values. 
For example, when $(d_1,d_2)=(+1,-1)$, $\ell_{+1,-1}(q,\sigma_1,\sigma_2)$ equals
\[
\mu_1(conf_{i}(uq\sigma \tau_2\tau_3 v\natural (r+1) \natural w),\delta(q,\sigma_1,\sigma_2)(p,\tau_1,\eta,+1,-1)),
\]
where $\sigma_2$ is the symbol at cell $r+1$.
Notice that each value  $\ell_{d_1,d_2}(q,\sigma_1,\sigma_2)$ is always defined.
If either $p\in F$ or $u'pv'\natural r\natural  w'$ has a wrong form, then we set $conf_{i+1}(u'pv'\natural r\natural w')=0$.

A {\em fuzzy computation} of $M$ on fuzzy input $s$ is a series $(conf_0,conf_1,\ldots)$ of fuzzy configurations defined above.
Let $final_{M}$
be an element of $\FF(\nat\times CONF_{M})$ defined as
\[
 final_{M}(t,uqv\natural r \natural  w) = [q\in F]\cdot \mu_3(\{conf_i(uqv\natural r \natural w)\}_{i\in[0,t]_{\integer}})
 \]
for any $uqv\natural r \natural w$ in $CONF_{M}$.
This value $final_{M}(t,uqv\natural r \natural w)$ is called the {\em possibility degree} of the (final) configuration
$uqv\natural r \natural w$ at time $t$. Let $FIN_{M}(t,s)$ be the set of all pairs $(uqv,r)$ with $q\in F$, $u,v\in\Sigma^*$, $|uv|\leq t$, and $r\in[0,\ell(s)]_{\integer}$. Obviously, $FIN_{M}(t,s)$ is a finite set.

Since we are interested in time-bounded computation, we define the {\em (absolute) running time} of $M$ on input $s$ to be a unique number $t\in\nat$ for which (i) there exists a final configuration $u'q'v'\natural r\natural w'$ satisfying $final_{M}(t,u'q'v'\natural r\natural w')>0$ and (ii)  $final_{M}(i,uqv\natural r\natural  w)=0$ holds for all configurations $uqv\natural r\natural w$ and all $i>t$.
For convenience, we say that {\em $M$ (absolutely) halts on input $s$ in time $t$} if $t$ is at least the running time of
$M$ on $s$.

Now, let $t$ be the running time of $M$ on $s$. If $b$ is a fuzzy subset of $\Gamma^*$ satisfying
\[
 b(w) = \xi(\{final_{M}(t,uqv\natural r\natural w)\}_{(uqv,r)\in FIN_{M}(t,s)})
\]
for every $w\in\Gamma^*$, then we say that $M$ {\em outputs $b$ on input $s$ in time $t$}. For convenience, we write $M(s)$ to denote this particular fuzzy object $b$.
Obviously, if $M$ outputs $b$ in time $t$, then $|\supp(b)|\leq |\Gamma|^{t}$ holds, and thus $b$ has a finite support.

We say that a fuzzy problem $L$ {\em admits} a DFTM $M$ (or $M$ solves $L$) if $M(s)=L(s)$ holds for any $s\in F\Theta_1$. In this case, the problem $L$ is succinctly denoted by $\LL(M)$.

\subsection{Fuzzy Gates and Fuzzy Circuits}

In an ordinary setting, uniform families of {\em Boolean circuits} composed of {\em Boolean gates} have been used to model efficient algorithms. Likewise,
we will consider a fuzzification of those circuits to model our intended fuzzy algorithms. In what follows, we fix a safe tuple  $\Xi'=(\mu_1,\mu'_2,\mu_3,\xi)$ of auxiliary operators.
A fuzzy gate is specified by a finite set of input variables attached with distinguished labels  and  another finite set of output variables having  specific labels.
A {\em fuzzy gate with $m$ inputs and $k$ outputs} simply represents a function mapping  $\{0,1\}^m$ to $\FF(\{0,1\}^k)$. For a physical implementation of fuzzy circuits, it may be desirable to restrict the usable types of fuzzy gates but, meanwhile, we allow any types of fuzzy gates.

In our fuzzy circuit model, a finite number of fuzzy gates are layered by ``levels.'' At each middle level, no two fuzzy gates must share output variables with the same label; however, fuzzy gates are allowed to share input variables with the same labels.
Nevertheless, all input variables at level $0$ have distinct labels.
We also permit the use of ``fuzzy constants'' (\ie elements in $\FF(\{0,1\}^{\ell})$ for a fixed $\ell$) as part of inputs.

Let us explain how a fuzzy circuit operates on a given set of inputs. Let $(x_1,x_2,\ldots,x_n)$ be a series of all input variables used for
a given fuzzy circuit.
We define a {\em global configuration} $conf_{t}$ at level $t$ as follows.
At level $0$, let $conf_{0}(x_1x_2\ldots x_n) = s(x_1x_2\cdots x_n)$. For  $t\geq 0$, let $G_1,G_2,\ldots,G_m$ be all fuzzy gates aligned at level $t+1$
and let each fuzzy gate $G_k$ have input variables $\vec{v}_k=(v_{i_1},v_{i_2},\ldots,v_{i_{a_k}})$ as well as output variables $\vec{w}_k=(w_{j_1},w_{j_2},\ldots,w_{j_{b_k}})$. Let $\vec{v}=(v_1,v_2,\ldots,v_{c})$ be the series of all input variables of $G_1,G_2,\ldots,G_m$ and let $(w_1,w_2,\ldots,w_{d})$ be the series of all output variables of them. Now, we define

\s
\n\hs{1}
$\bullet$\hs{2} $\ell^{(k)}_{\vec{v}'}(\vec{v}'_k,\vec{w}'_k) = \mu_1(conf_{t}(v'_1,\ldots,v'_c),G_k(\vec{v}'_{k})
(\vec{w}'_k))$, and \\
\n\hs{1}
$\bullet$\hs{2} $g(v'_1,\ldots,v'_c)(w'_1,\ldots,w'_d) = \mu'_1(\{\ell^{(k)}_{\vec{v}'}(\vec{v}'_k,\vec{w}'_k)\}_{k\in[m]})$,
\s

\n where, given a variable $a$, $a'$ means its arbitrary value.

Finally, we define $conf_{t+1}(w'_1,\ldots,w'_d)$ to be
\[
\mu_3(\{g(v'_1,\ldots,v'_c)(w'_1,\ldots,w'_d)\}_{(v'_1,\ldots,v'_c)\in\supp(conf_t)}).
\]
An {\em output} (or {\em outcome}) of $C$ on input $s$ is $conf_t$ when $t$ is the number of levels in $C$ and we write $C(s)$ to denote the outcome of $C$ on $s$. Moreover, the {\em size} of $C$ is the total number of fuzzy gates plus the number of all wires between variables and gates used in $C$.

Now, we consider the aforementioned standard safe tuple of auxiliary operators together with $\mu'_2(\{\alpha_r\}_{r\in B}) = \bigwedge_{r\in B} \alpha_r$.

\begin{theorem}\label{circuit-DFTM-translate}
Let $f$ be any fuzzy function from $F\Theta_1$ to $F\Theta_2$.

\n\hs{1}
(1) If there is a DFTM computing $f$ in polynomial time, then there exists a uniform family of fuzzy circuits of polynomial size that computes $f$. \\
\n\hs{1}
(2) If there is a uniform family of polynomial-size fuzzy circuits computing $f$, then there exists a DFTM $M$ that computes $f$ in polynomial time.
\end{theorem}

\vs{-2}
\begin{proofsketch}
We prove only (1).  Let $M=(Q,\Delta,\Sigma_1,\Sigma_2,\Gamma,\delta,q_0,F)$ be a DFTM computing $f$ in polynomial time. We want to construct a uniform family $\{C_n\}_{n\in\nat}$ of fuzzy circuits that ``simulates'' $M$, and thus computes $f$.
For each fuzzy gate $g_k$ at level $t$, we prepare variables $\{v_{t,\sigma,k},v'_{t,q,k}\}_{t,\sigma,q,k}$, where
$t\in\nat$, $q\in\Delta\cup\{\cent,\dollar\}$, $q\in Q$, and $k\in\integer$, used for inputs and outputs.
Intuitively, $v_{t,\sigma,k}=v'_{t,q,k}=1$, for example, means that $M$ in state $q$ is scanning $\sigma$ on cell $k$ at time $t$.
We assume a natural and efficient ordering of those  variables.
It is easy to design a fuzzy gate that ``mimics'' the behavior of $\delta(q,\sigma_1,\sigma_2)$. Since $M$ halts in time polynomial in the length $\ell(s)$ of $s$, the size of $C_n$ is also upper-bounded by a certain polynomial in $\ell(s)$. However, $g_{k-1},g_k,g_{k+1}$ may share the same output variables. To avoid this situation, it is enough to distinguish them by  slightly modifying their labels.
\end{proofsketch}

\subsection{Fuzzy Proof Verification Systems}\label{sec:FPVS}

Nondeterminism has played a significant role in the development of the
theory of NP-completeness.
In the past literature, the notion of polynomial time-bounded nondeterminism has been characterized
in various ways. Among them, we are interested in a particular characterization of NP problems, using proof verification processes; namely, NP problems are decision problems that have efficiently verifiable solutions (or proofs).
Likewise, we will introduce a model of {\em fuzzy proof verification system}, in which a verifier tries to check the validity (or correctness) of a given proof, because we believe that this model is more suitable for practical use.

Let us recall the definition of DFTM from Section \ref{sec:DFTM} and expand it significantly by adding a mechanism of handling proofs (or certificates).
To store such a proof during a verification process, we use an extra read-only infinite tape, called a {\em proof tape}.  A {\em fuzzy proof verification system} (or an FPVS, in short) is a tuple $\pair{N,F\Theta,\Xi,I}$, where $F\Theta$ is a set of all possible fuzzy proofs with $F\Theta\subseteq \FF(\Delta_2^*)$ and $N = (Q,\Delta_1,\Delta_2, \Sigma_1,\Sigma_2,\Sigma_3, \Gamma,\delta,q_0,F)$ is a 4-tape DFTM whose first and second tapes are the same as before, whose third  tape is a read-only proof tape, and whose fourth one is a write-only output tape. Here, a fuzzy transition function $\delta$ is a map from $\delta:(Q-F)\times \Sigma_1\times \Sigma_2 \times \Sigma_3$ to $\FF(Q\times \Sigma_1\times \Sigma_2\times \Gamma\times D_1\times D_2\times D_3)$, where the $i$th tape uses a tape alphabet $\Sigma_i$ and its tape head moves in directions specified as $D_i$ ($=\{0,\pm1\}$) for each index $i\in [3]$.
A configuration is now of the form $upv\natural r \natural y_1hy_2\natural w$, where $y=y_1y_2$ particularly represents the content of the proof tape and $y_1hy_2$ indicates that its tape head is at the leftmost symbol of $y_2$. On input $s$ with proof $\phi$, a fuzzy configuration $conf_i$ at time $i$ is defined as follows. Let $ conf_0(upv\natural  r \natural y_1 hy_2\natural w)$ denote
\[
 s(x)\cdot \phi(y)\cdot [upv\natural  r \natural y_1hy_2\natural w = q_0\cent x\dollar \natural 0 \natural hy \natural \lambda].
\]
At time $i+1$, we assume that $u'pv'\natural  r \natural y'_1hy'_2\natural w'$ is of the form $u\tau_1p\tau_2\tau_3\natural  r \natural y_1\xi_1 h\xi_2\xi_3 y_2\natural w\eta$.
Let $conf_{i+1}(u'pv'\natural  r \natural y'_1hy'_2\natural w')$ denote
\[
 \mu_2(\{\ell_{d_1,d_2,d_3} (q,\sigma_1,\sigma_2,\sigma_3)\}_{(q,\sigma_1,\sigma_2,\sigma_3), (d_1,d_2,d_3)}),
\]
where $(q,\sigma_1,\sigma_2,\sigma_3)\in (Q-F)\times \Sigma_1\times \Sigma_2\times \Sigma_3$, $d_1,d_2,d_3\in \{0,\pm1\}$,  each $\ell_{d_1,d_2,d_3}(q,\sigma_1,\sigma_2,\sigma_3)$ is defined similarly to $\ell_{d_1,d_2}(q,\sigma_1,\sigma_2)$ in Section \ref{sec:DFTM}; for example, $\ell_{+1,-1,+1}(q,\sigma_1,\sigma_2,\sigma_3)$ equals
\[
\mu_1(conf_{i}(u'\natural r' \natural y'\natural w, \delta(q,\sigma_1,\sigma_2,\sigma_3)(p,\tau_1,\eta,+1,-1,+1)),
\]
where $r'=r+1$ and $u'\natural r'\natural y'\natural w$ expresses the string $uq\sigma_1 \tau_2\tau_3 v\natural r'\natural y_1h\xi_1\xi_2\xi_3\natural w$.

Let $FIN_{N}(t,s,\phi)$ be the set of all $(uqv,r,y_1hy_2)$ with $u,v\in\Sigma_1^*$, $y_1y_2\in\supp(\phi)$, $q\in F$, $|uv|\leq t$, and $r\in[0,\ell(s)]_{\integer}$. As in Section \ref{sec:DFTM}, $final_{N}$ and the running time $t$ of $N$ are  similarly defined. Write $N(s,\phi)$ to denote an output (\ie a fuzzy subset of $\{0,1\}$) $b$ of $N$ on input $s$ with proof $\phi$, where $b(w)$ is defined as
\[
 \xi(\{final_{N}(t,uqv\natural r\natural y_1hy_2\natural w)\}_{(uqv,r,y_1hy_2)\in FIN_{N}(t,s,\phi)}).
\]
An {\em outcome} $N(s)$ of the FPVS $N$ on input $s$ is defined as

\n\hs{1}$\bullet$\hs{2}  $N(s)(1) = \sup_{\phi\in F\Theta}\{N(s,\phi)(1)\}$
and \\
\n\hs{1}$\bullet$\hs{2}  $N(s)(0) = \inf_{\phi\in F\Theta}\{N(s,\phi)(0)\}$.

We say that $L$ {\em admits} $N$ (or $N$ {\em solves} $L$) if $N(s)=L(s)$ for all $s\in F\Theta_1$. The notation $\LL(N)$ expresses the fuzzy decision problem solved by $N$.


As a concrete example of fuzzy decision problems admitting FPVS's, we present the Fuzzy Circuit Satisfiability Problem (abbreviated as Fuzzy-Circuit-SAT).

\s
\n\hs{1}{\sc Fuzzy-Circuit-SAT (w.r.t. $F\Theta$)}

\n\hs{1}$\bullet$\hs{2} {\sc Fuzzy Instance:} a (description of) fuzzy circuit $C$  that takes inputs from $F\Theta$. \\
\n\hs{1}$\bullet$\hs{2} {\sc Fuzzy Output:} output a fuzzy set $b\in \FF(\{0,1\})$, where $b(1) = \sup_{s\in F\Theta}\{C(s)(1)\}$ and $b(0) = \inf_{s\in F\Theta}\{C(s)(0)\}$.

\s

To solve this Fuzzy-Circuit-SAT, it is possible to construct an
FPVS using Theorem \ref{circuit-DFTM-translate}.
Hence, we obtain the following.

\begin{lemma}\label{FPVS-Circuit-SAT}
There exists an FPVS that solves $Fuzzy\mbox{-}Circuit\mbox{-}SAT$ in polynomial time.
\end{lemma}

\section{Reductions among Fuzzy Problems}\label{sec:approximation}

The notion of {\em reducibility} is a basis to NP-completeness and numerous forms of the reducibility have been proposed. Here, we will consider only its  simple fuzzification.
Firstly, we will define an important fuzzy function class $\fuzzyfpa(\gamma)$ of polynomial-time approximately computable fuzzy functions.
Using functions in $\fuzzyfpa(\gamma)$, we will introduce AF- and APF-reducibilities, which are viewed as natural fuzzifications of Krentel's metric reducibility \cite{Kre88} and AP-reducibility \cite{ACG+03}.

\subsection{Relative Approximation of Membership Degree}\label{sec:membership-degree}

In Section \ref{sec:DFTM}, we have introduced a DFTM model to capture the notion of fuzzy algorithm. However, the exact use of DFTM's seems too restrictive to solve a wide range of practical fuzzy problems. It is thus  desirable to allow the DFTM's to ``approximate'' the outcomes of the fuzzy problems.

To describe this ``approximation,'' let $\gamma:\nat\rightarrow[1,\infty)$ be any crisp function, called an {\em imprecision tolerance parameter}.
Let $F\Theta\subseteq F\Delta^*$ for alphabet $\Delta$ as before.
For any two fuzzy subsets $F$ and $G$ of the universe $F\Theta$, we say that $F$ is a {\em  $\gamma$-approximation} of $G$ if
$
F(x)/{\gamma(|x|)} \leq {G(x)}\leq \gamma(|x|)F(x)
$
for any element $x$ in $F\Theta$.

\begin{definition}[approximate solving]
Let $M$ be either an DFTM or an FPVS. We say that $M$ {\em $\gamma$-approximately solves} fuzzy problem $L$ if, for any admissible fuzzy input $s$, $M$ takes $s$ as an input and produces a certain fuzzy solution $M(s)$ that is a $\gamma$-approximation of $L(s)$.
\end{definition}

In ordinary computational complexity theory, $\fp$ denotes the set of all crisp functions on crisp strings computable in polynomial time. We introduce an analogous ``fuzzy'' function class denoted by $\fuzzyfpa(\gamma)$ using the $\gamma$-approximability.

\begin{definition}[Fuzzy-FPA$(\gamma)$]
Let $\gamma:\nat\rightarrow[1,\infty)$ be any imprecision tolerance parameter and fix a safe tuple $\Xi$.
For convenience, we define $\fuzzyfpa_{\Xi}(\gamma)$ to be the set of all (combinatorial generic) fuzzy problems, each of which
can be $\gamma$-approximately solved by a certain polynomial-time DFTM. The ``PA'' stands for ``polynomial-time approximate.''  When $\Xi$ is clear from the context, we drop $\Xi$ and write $\fuzzyfpa(\gamma)$.
\end{definition}

Let us introduce four classes  of fuzzy functions. The notation $const$ expresses the set of all constant functions (which we usually identify with ``constants''), $poly$ does the set of all polynomials, and $exp$ does the set of all exponential functions.
We then obtain the following chain of containments:
$
\fuzzyfpa(1) \subseteq \fuzzyfpa(const)\subseteq \fuzzyfpa(poly)\subseteq \fuzzyfpa(exp).
$

We can show that $\fp$ is a crisp part of $\fuzzyfpa$ if we take
the following method of fuzzifying crisp objects.
For each crisp string $x$ in $\Delta^*$,
the notation $\hat{x}$ denotes its specific fuzzified object $\{(x,1)\}\cup\{(y,0)\mid y\in \Delta^*-\{x\}\}$ in $\FF(\Delta^*)$.

\begin{lemma}
For any crisp function $f:\Delta^*\to\Gamma^*$ in $\fp$, there exists a fuzzy function  $g\in \fuzzyfpa(1)$ such that, for every $(x,y)\in\Delta^*\times \Gamma^*$, $f(x)=y$ if and only if $g(\hat{x}) = \hat{y}$.
\end{lemma}

\begin{lemma}
Let $\FF\in\{1,const,poly,exp\}$. $\fuzzyfpa(\FF)$ is closed under functional composition; namely, for any two elements $f,g\in\fuzzyfpa(\FF)$, the function $h$ defined as $h(s)=g(f(s))$ for every $s$ is also in $\fuzzyfpa(\FF)$.
\end{lemma}

\subsection{Approximate Fuzzy Reductions}

To compare the computational complexities of two (generic) fuzzy problems, we need to devise a notion of ``fuzzy reducibility,''
which is a mechanism (similar to the metric reducibility in \cite{Kre88}) of transforming instances of one decision problem to instances of another decision problem so that corresponding outputs
of those instances are close enough.

\begin{definition}[AF-reducibility]
Let $F$ be any fuzzy function mapping $F\Theta_1$ to $F\Theta_2$ and let $G$ be any fuzzy function from $F\Theta_3$ to $F\Theta_4$. We say that $F$ is {\em polynomial-time $\gamma$-approximately fuzzy reducible} (or {\em AF$(\gamma)$-reducible}, in short) to $G$ if there exist two functions $f,g$ in $\fuzzyfpa(\FF)$ with $f:F\Theta_1\rightarrow F\Theta_3$ and $g:F\Theta_1\times F\Theta_4\rightarrow F\Theta_2$ and a function $\gamma\in \FF$ such that, for any $s\in F\Theta_1$, $g(s,G(f(s)))$ is a $\gamma$-approximation of $F(s)$. In this case, we write $F\AFreduces^{(\FF)} G$. This triplet $(f,g,\gamma)$ is called an {\em AF$(\FF)$-reduction} of $F$ to $G$.
\end{definition}

The next lemma establishes the reflexivity and the transitivity of the AF-reducibility. Thus, the AF$(\FF)$-reducibility forms a partial order in the set of all fuzzy problems.

\begin{lemma}
Let $\FF\in\{1,const,poly,exp\}$. For any three fuzzy problems $A,B,C$ having the same range, it holds that (1) $A\AFreduces^{(\FF)} A$ and  (2) $A\AFreduces^{(\FF)} B$ and $B\AFreduces^{(\FF)} C$ imply $A\AFreduces^{(\FF)} C$.
\end{lemma}

Reducibility between two fuzzy optimization problems, in contrast, requires a more delicate treatment. As a natural fuzzification of the ordinary AP-reducibility (see, \eg \cite{ACG+03}), we introduce the following APF-reducibility. Let $R(s,t)= \max\{\left| m(s,u)/m^*(s)\right|, \left| m^*(s)/m(s,u)\right|\}$ for a given fuzzy measure function $m$.
We set $\rational^{>1}=\{r\in \rational\mid r>1\}$.

\begin{definition}[APF-reducibility]
For any two fuzzy optimization problems $A=(I_1,SOL_1,m_1,goal)$ and $B=(I_2,SOL_2,m_2,goal)$, we say that $A$ is
{\em polynomial-time approximation-preserving fuzzy (APF) reducible to} $B$ with respect to $\FF$, denoted by  $A\APFreduces^{(\FF)} B$, if the following condition holds: there exist a pair $(f,g)$ of fuzzy functions and a constant $c>0$ such that

\n\hs{1}
(1)\hs{1} $f(s,r)\in I_2$ for any $s\in I_1$ and $r\in\rational^{>1}$; \\
\n\hs{1}
(2)\hs{1} $SOL_1(s)\neq\setempty$ implies $SOL_2(f(s,r))\neq\setempty$; \\
\n\hs{1}
(3)\hs{1} $u\in SOL_2(f(s,r))$ implies $g(s,u,r)\in SOL_1(s)$; \\
\n\hs{1}
(4)\hs{1} $f,g\in\fuzzyfpa(\FF)$ for each fixed $r\in\rational^{>1}$; \\
\n\hs{1}
(5)\hs{1} $R_2(f(s,r),u)\leq r$ $\rightarrow$ $R_1(s,g(s,u,r))\leq 1+c(r-1)$.
\end{definition}

\section{Fuzzy Decision Problems}

In ordinary complexity theory, decision problems are identified with sets of words (or strings) and they are also called languages. The fundamental complexity classes of languages are P and NP. Analogously, we  recognize two special classes of fuzzy decision problems.

\begin{definition}[Fuzzy-PA$(\gamma)$, Fuzzy-NPA$(\gamma)$]
Let $\fuzzypa(\gamma)$ be a subclass of $\fuzzyfpa(\gamma)$, which consists only of fuzzy decision problems. Moreover, $\fuzzynpa(\gamma)$ denotes the set of all fuzzy decision problems that can be $\gamma$-approximately solved by polynomial-time FPVS's.
\end{definition}

Since DFTMs are a special case of FPVS's, it immediately follows that
$\fuzzypa(\gamma)\subseteq \fuzzynpa(\gamma)$ for any $\gamma$. As in Section \ref{sec:membership-degree}, we write $\fuzzypa$ (resp., $\fuzzynpa$) for $\fuzzypa(const)$ (resp., $\fuzzynpa(const)$).

Notice that DFTMs are in fact an extension of ordinary NTMs. Therefore, it is not surprising to show that all $\np$ sets can be solved by DFTMs in polynomial time if we use an appropriate fuzzification (\ie if we assign appropriately-chosen possibility degrees).
Recall from Section \ref{sec:membership-degree} the fuzzification $\hat{x}$ of crisp string $x$. If a fuzzification $\hat{L}$ of crisp language $L$ is defined to satisfy that $\hat{L}(\hat{x}) =\{(0,0),(1,1)\}$ if $x\in L$ and $\hat{L}(\hat{x}) =\{(0,1),(1,0)\}$ otherwise, we can show that, for any language $L\in \np$, $\hat{L}$ belongs to $\fuzzypa(1)$.

\begin{proposition}\label{NP-vs-Fuzzy-PA}
For every set $A\in\np$ over alphabet $\Delta$,
there exists a fuzzy decision problem $B\in\fuzzypa(1)$ such that $A=\{x\in\Delta^*\mid B(\hat{x})(1) =1\}$ and $\overline{A}=\{x\in\Delta^*\mid B(\hat{x})(1) = 0\}$.
\end{proposition}


Hereafter, we assume that $\FF\in\{1,const,poly,exp\}$.

\begin{lemma}\label{FuzzyPA-closure}
The fuzzy complexity class $\fuzzypa(\FF)$ is closed under AF$(\FF)$-reductions.
\end{lemma}

\vs{-2}
\begin{proofsketch}
It suffices to show that, for any two fuzzy problems $A$ and $B$, if $A$ is AP-reducible to $B$ and $B$ is in $\fuzzypa$, then $A$ is also in $\fuzzypa(\FF)$.  Since $A\AFreduces^{(\FF)} B$, take a reduction $(f,g,\gamma)$ with $f,g\in\fuzzyfpa(\FF)$ that reduces $A$ to $B$. Since $B\in\fuzzypa(\FF)$, take a DFTM $M$ that solves $B$. It suffices to consider the following DFTM $G$: on input $s$, generate all $x\in\supp(s)$ in parallel, compute $M_f$ on them, starts the simulation of $N$, and compute $M_g$.
\end{proofsketch}

Now, we introduce a key concept of ``completeness.''

\begin{definition}[completeness]
We say that a fuzzy decision problem $A$ is {\em complete} for $\fuzzynpa(\FF)$ (or simply, $\fuzzynpa(\FF)$-complete) if (i)  $A$ is in $\fuzzynpa(\FF)$ and (ii) for every fuzzy decision problem $B$ in $\fuzzynpa(\FF)$, $A$ is AF$(\FF)$-reducible to $B$.
\end{definition}

\begin{lemma}
Let $A$ be a fuzzy decision problem complete for $\fuzzynpa(\FF)$. If $A$ is in $\fuzzypa(\FF)$, then $\fuzzypa(\FF) =\fuzzynpa(\FF)$ holds.
\end{lemma}

\vs{-2}
\begin{proofsketch}
The containment $\fuzzypa\subseteq \fuzzynpa$ is obvious. Next, we will show the other containment. Take any fuzzy problem $B$ in $\fuzzynpa$. By the definition of ``completeness,'' $B$ is AP-reducible to $A$. Assume that $A$ is in $\fuzzypa$. By Lemma \ref{FuzzyPA-closure}, it follows that $B$ is also in $\fuzzypa$. Thus, $\fuzzynpa\subseteq \fuzzypa$.
\end{proofsketch}

Finally, we will demonstrate the existence of complete problems for $\fuzzynpa(1)$. Our choice of such problem is Fuzzy-Circuit-SAT, defined in Section \ref{sec:FPVS}.

\begin{theorem}
The fuzzy problem Fuzzy-Circuit-SAT is complete for $\fuzzynpa(1)$.
\end{theorem}

\vs{-2}
\begin{proofsketch}
Recall from Lemma \ref{FPVS-Circuit-SAT} that $Fuzzy\mbox{-}Circuit\mbox{-}SAT$ is in $\fuzzynpa(1)$. What remains to show is that any problem, say, $A$ in $\fuzzynpa(1)$ is AF-reducible to $Fuzzy\mbox{-}Circuit\mbox{-}SAT$. Given such $A$, take an FPVS $N$ for $A$. By Theorem \ref{circuit-DFTM-translate}(1), we can choose a family of fuzzy circuits that ``simulates'' $N$ when input $s$ and proof $\phi$ are initially given. It thus suffices to define an AF-reduction pair $(f,g)$ as follows. Let $f(s)$ be a circuit obtained from $C$ by incorporating it with $s$ (treating $\phi$ as only a true input). The function $g$ is defined as the identity function.
\end{proofsketch}

\section{Fuzzy Optimization Problems}

Briefly, we will discuss the computational complexity of
fuzzy optimization problems. The theory of NP optimization problems has made a huge success in classifying ``complete'' problems in NPO (the class of NP optimization problems). We  will pay our attention to ``fuzzy NPA'' optimization problems.

\begin{definition}[Fuzzy-NPO$(\gamma)$]
A {\em fuzzy NPA$(\gamma)$ optimization problem} (or a fuzzy NPAO$(\gamma)$ problem) $P$ is a fuzzy optimization problem $(I,SOL,m,goal)$ as defined in Section \ref{sec:fuzzy-problem} such that $I\circ SOL$ is in $\fuzzypa(\gamma)$ and $m$ is in $\fuzzyfpa(\gamma)$. We write $\fuzzynpao(\gamma)$ for the class of all fuzzy NPAO$(\gamma)$ problems.
\end{definition}

Similar to $\fuzzynpa(1)$, $\fuzzynpao(1)$ has complete problems under AFP$(1)$-reductions.

\begin{theorem}
There exists a fuzzy optimization problem that is complete for $\fuzzynpao(1)$ under APF$(1)$-reductions.
\end{theorem}

\let\oldbibliography\thebibliography
\renewcommand{\thebibliography}[1]{%
  \oldbibliography{#1}%
  \setlength{\itemsep}{0pt}%
}
\bibliographystyle{plain}

\end{document}